\documentclass[final]{IEEEtran}

\usepackage[pdftex]{graphicx}

\usepackage{amssymb}
\usepackage{amsmath}

\usepackage[section]{algorithm}
\usepackage{algorithmic}

\usepackage{bm}
\newcommand{\Trsps}{^{\mathrm{T}}}

\newcommand{\Fig}[1]{Fig.\ \ref{fig:#1}}
\newcommand{\Eq}[1]{(\ref{eq:#1})}
\newcommand{\Table}[1]{Table\ \ref{tbl:#1}}

\graphicspath{{../pics/}}
\usepackage{epstopdf}

\usepackage{subfigure}

\begin{document}
%
\title{Learning Domain-Invariant Subspace using \\ Domain Features and Independence Maximization}

\author{Ke~Yan,
        Lu Kou,
        and David~Zhang,~\IEEEmembership{Fellow,~IEEE}
\thanks{The work is partially supported by the GRF fund from the HKSAR Government, the central fund from Hong Kong Polytechnic University, the NSFC fund (61332011, 61272292, 61271344), Shenzhen Fundamental Research fund (JCYJ20150403161923528, JCYJ20140508160910917), and Key Laboratory of Network Oriented Intelligent Computation, Shenzhen, China.}
\thanks{K. Yan is with the Department of Electronic Engineering, Graduate School
	at Shenzhen, Tsinghua University, Shenzhen 518055, China (e-mail:
	yankethu@foxmail.com).}
\thanks{L. Kou is with the Department of Computing, The Hong Kong Polytechnic University, Kowloon, Hong Kong (e-mail: cslkou@comp.polyu.edu.hk).}
\thanks{D. Zhang is with the Shenzhen Graduate School, Harbin Institute of Technology, Shenzhen 518055, China, and also with the Department of Computing, Biometrics Research Centre, The Hong Kong Polytechnic University, Kowloon, Hong Kong (e-mail: csdzhang@comp.polyu.edu.hk).}}

\markboth{Accepted by IEEE TRANSACTIONS ON CYBERNETICS}%
{Accepted by IEEE TRANSACTIONS ON CYBERNETICS}

\maketitle

\begin{abstract}
	
Domain adaptation algorithms are useful when the distributions of the training and the test data are different. In this paper, we focus on the problem of instrumental variation and time-varying drift in the field of sensors and measurement, which can be viewed as discrete and continuous distributional change in the feature space. We propose maximum independence domain adaptation (MIDA) and semi-supervised MIDA (SMIDA) to address this problem. Domain features are first defined to describe the background information of a sample, such as the device label and acquisition time. Then, MIDA learns a subspace which has maximum independence with the domain features, so as to reduce the inter-domain discrepancy in distributions. A feature augmentation strategy is also designed to project samples according to their backgrounds so as to improve the adaptation. The proposed algorithms are flexible and fast. Their effectiveness is verified by experiments on synthetic datasets and four real-world ones on sensors, measurement, and computer vision. They can greatly enhance the practicability of sensor systems, as well as extend the application scope of existing domain adaptation algorithms by uniformly handling different kinds of distributional change.

\end{abstract}


%
\IEEEpeerreviewmaketitle

\section{Introduction} \label{sec:intro}

\IEEEPARstart{I}{n} many real-world machine learning problems, the labeled training data are from a source domain and the test ones are from a target domain. Samples of the two domains are collected under different conditions, thus have different distributions. Labeling samples in the target domain to develop new prediction models is often labor-intensive and time-consuming. Therefore, domain adaptation or transfer learning is needed to improve the performance in the target domain by leveraging unlabeled (and maybe a few labeled) target samples \cite{Pan10TransSurvey}. This topic is receiving increasing attention in recent years due to its broad applications such as computer vision \cite{Patel15VDASurvey, Cui14Flow, Bian12ActRec} and text classification \cite{Pan11TCA, Seah13Combat}. It is also important in the field of sensors and measurement. Because of the variations in the fabrication of sensors and devices, the responses to the same signal source may not be identical for different instruments, which is known as instrumental variation. Furthermore, the sensing characteristics of the sensors, the operating condition, or even the signal source itself, can change over time, which leads to complex time-varying drift. As a result, the prediction model trained with the samples from the initial device in an earlier time period (source domain) is not suitable for new devices or in a latter time (target domains).

A typical application plagued by this problem is machine olfaction, which uses electronic noses (e-noses) and pattern recognition algorithms to predict the type and concentration of odors \cite{Gardner94History}. The applications of machine olfaction range from agriculture and food to environmental monitoring, robotics, biometrics, and disease analysis \cite{Rock08Review, Marj14TCybGas, ZhLei11OLTrans, Yan14System}. However, owing to the nature of chemical sensors, many e-noses are prone to instrumental variation and time-varying drift mentioned above \cite{Marco12Review, Dicarlo12Drift}, which greatly hamper their usage in real-world applications. Traditional methods dealing with these two kinds of drift (``drift correction'' methods hereinafter) require a set of transfer samples, which are predefined gas samples needed to be collected with each device and in each time period \cite{Marco12Review, ZhLei11OLTrans, Yan15WpdsSemi, Yan16TCTL}. They are often used to learn regression models to map the features in the target domain to the source domain \cite{ZhLei11OLTrans, Yan15WpdsSemi}. Nevertheless, collecting transfer samples repeatedly is a demanding job especially for non-professional e-nose users.

In such cases, domain adaptation techniques with unlabeled target samples are desirable. An intuitive idea is to reduce the inter-domain discrepancy in the feature level, i.e.\ to learn domain-invariant feature representation \cite{Pan11TCA, Shi12Itl, Fern13Sa, Cui14Flow, Gong14Gfk, Shao14Ltsl, Blit07SCLMI, Chen12mSDA, Jiang16IGLDA}. For example, Pan et al.\ \cite{Pan11TCA} proposed transfer component analysis (TCA), which finds a latent feature space that minimizes the distributional difference of two domains in the sense of maximum mean discrepancy. More related methods will be introduced in Section \ref{subsec:unsupDa}. When applied to drift correction, however, existing domain adaptation algorithms are faced with two difficulties. First, they are designed to handle discrete source and target domains. In time-varying drift, however, samples come in a stream, so the change in data distribution is often continuous. One solution is to split data into several batches, but it will lose the temporal order information. Second, because of the variation in the sensitivity of chemical sensors, the same signal in different conditions may indicate different concepts. In other words, the conditional probability $ P(Y|X) $ may change for samples with different backgrounds, where ``background'' means when and with which device a sample was collected. Methods like TCA project all samples to a common subspace, hence the samples with similar appearance but different concepts cannot be distinguished.

In this paper, we present a simple yet effective algorithm called maximum independence domain adaptation (MIDA). The algorithm first defines ``domain features'' for each sample to describe its background. Then, it finds a latent feature space in which the samples and their domain features are maximally independent in the sense of Hilbert-Schmidt independence criterion (HSIC) \cite{Gretton05Hsic}. Thus, the discrete and continuous change in distribution can be handled uniformly. In order to project samples according to their backgrounds, feature augmentation is performed by concatenating the original feature vector with the domain features. We also propose semi-supervised MIDA (SMIDA) to exploit the label information with HSIC. MIDA and SMIDA are both very flexible. (1) They can be applied in situations with single or multiple source or target domains thanks to the use of domain features. In fact, the notion ``domain'' has been extended to ``background'' which is more informative. (2) Although they are designed for unsupervised domain adaptation problems (no labeled sample in target domains), the proposed methods naturally allow both unlabeled and labeled samples in any domains, thus can be applied in semi-supervised (both unlabeled and labeled samples in target domains) and supervised (only labeled samples in target domains) problems as well. (3) The label information can be either discrete (binary- or multi-class classification) or continuous (regression).

To illustrate the effect of our algorithms, we first evaluate them on several synthetic datasets. Then, drift correction experiments are performed on two e-nose datasets and one spectroscopy dataset. Note that spectrometers suffer the same instrumental variation problem as e-noses \cite{Feud02TransRev}. Finally, a domain adaptation experiment is conducted on a well-known object recognition benchmark: Office+Caltech \cite{Gong12Gfk}. Results confirm the effectiveness of the proposed algorithms. The rest of the paper is organized as follows. Related work on unsupervised domain adaptation and HSIC is briefly reviewed in Section \ref{sec:relWork}. Section \ref{sec:method} describes domain features, MIDA, and SMIDA in detail. The experimental configurations and results are presented in Section \ref{sec:exp}, along with some discussions. Section \ref{sec:conclusion} concludes the paper.

\section{Related Work} \label{sec:relWork}

\subsection{Unsupervised Domain Adaptation} \label{subsec:unsupDa}

Two good surveys on domain adaptation can be found in \cite{Pan10TransSurvey} and \cite{Patel15VDASurvey}. In this section, we focus on typical methods that extract domain-invariant features. In order to reduce the inter-domain discrepancy while preserving useful information, researchers have developed many strategies. Some algorithms project all samples to a common latent space \cite{Pan11TCA, Shi12Itl, Shao14Ltsl}. Transfer component analysis (TCA) \cite{Pan11TCA} tries to learn transfer components across domains in a reproducing kernel Hilbert space (RKHS) using maximum mean discrepancy. It is further extended to semi-supervised TCA (SSTCA) to encode label information and preserve local geometry of the manifold. Shi et al.\ \cite{Shi12Itl} measured domain difference by the mutual information between all samples and their binary domain labels, which can be viewed as a primitive version of the domain features used in this paper. They also minimized the negated mutual information between the target samples and their cluster labels to reduce the expected classification error. The low-rank transfer subspace learning (LTSL) algorithm presented in \cite{Shao14Ltsl} is a reconstruction guided knowledge transfer method. It aligns source and target data by representing each target sample with some local combination of source samples in the projected subspace. The label and geometry information can be retained by embedding different subspace learning methods into LTSL.

Another class of methods first project the source and the target data into separate subspaces, and then build connections between them \cite{Fern13Sa, Gong12Gfk, Liu14Semi, Cui14Flow}. Fernando et al. \cite{Fern13Sa} utilized a transformation matrix to map the source subspace to the target one, where a subspace was represented by eigenvectors of PCA. The geodesic flow kernel (GFK) method \cite{Gong12Gfk} measures the geometric distance between two different domains in a Grassmann manifold by constructing a geodesic flow. An infinite number of subspaces are combined along the flow in order to model a smooth change from the source to the target domain. Liu et al.\ \cite{Liu14Semi} adapted GFK to correct time-varying drift of e-noses. A sample stream is first split into batches according to the acquisition time. The first and the latest batches (domains) are then connected through every intermediate batch using GFK. Another improvement of GFK is domain adaptation by shifting covariance (DASC) \cite{Cui14Flow}. Observing that modeling one domain as a subspace is not sufficient to represent the difference of distributions, DASC characterizes domains as covariance matrices and interpolates them along the geodesic to bridge the domains.

\subsection{Hilbert-Schmidt Independence Criterion (HSIC)} \label{subsec:hsic}

HSIC is used as a convenient method to measure the dependence between two sample sets $ X $ and $ Y $. Let $ k_x $ and $ k_y $ be two kernel functions associated with RKHSs $ \mathcal{F} $ and $ \mathcal{G} $, respectively. $ p_{xy} $ is the joint distribution. HSIC is defined as the square of the Hilbert-Schmidt norm of the cross-covariance operator $ \mathcal{C}_{xy} $ \cite{Gretton05Hsic}:
\begin{align*} \label{eq:hsicDef}
	&\text{HSIC}(p_{xy},\mathcal{F},\mathcal{G}) = \|\mathcal{C}_{xy}\|_{\text{HS}}^2 \\
		= &\mathbf{E}_{xx'yy'}[k_x(x,x')k_y(y,y')] + \mathbf{E}_{xx'}[k_x(x,x')]\mathbf{E}_{yy'}[k_y(y,y')] \\
		&- 2\mathbf{E}_{xy}[\mathbf{E}_{x'}[k_x(x,x')]\mathbf{E}_{y'}[k_y(y,y')]].
\end{align*}
Here $ \mathbf{E}_{xx'yy'} $ is the expectation over independent pairs $ (x, y) $ and $ (x', y') $ drawn from $ p_{xy} $. It can be proved that with characteristic kernels $k_x$ and $k_y$, HSIC$ (p_{xy},\mathcal{F},\mathcal{G}) $ is zero if and only if $ x $ and $ y $ are independent \cite{Song12FtSelHsic}. A large HSIC suggests strong dependence with respect to the choice of kernels. HSIC has a biased empirical estimate. Suppose $ Z = X \times Y = \{(x_1,y_1), \ldots, (x_n,y_n)\} $, $ K_x, K_y \in {\bf{R}}^{n\times n}$ are the kernel matrices of $ X $ and $ Y $, respectively, then \cite{Gretton05Hsic}:
\begin{equation} \label{eq:hsicEmp}
	\text{HSIC}(Z,\mathcal{F},\mathcal{G}) = (n-1)^{-2} \text{tr}(K_xHK_yH),
\end{equation}
where $ H = I-n^{-1}\bm{1}_n\bm{1}_n\Trsps \in {\bf{R}}^{n\times n}$ is the centering matrix.

Due to its simplicity and power, HSIC has been adopted for feature extraction \cite{Song07Cmvu, Pan11TCA, Bar11Spca} and feature selection \cite{Song12FtSelHsic}. Researchers typically use it to maximize the dependence between the extracted/selected features and the label. However, to our knowledge, it has not been utilized in domain adaptation to reduce the dependence between the extracted features and the domain features.

\section{Proposed Method} \label{sec:method}

\subsection{Domain Feature} \label{subsec:domainFt}

We aim to reduce the dependence between the extracted features and the background information. A sample's background information should (1) naturally exist, thus can be easily obtained; (2) have different distributions in training and test samples; (3) correlate with the distribution of the original features. The domain label (which domain a sample belongs) in common domain adaptation problems is an example of such information. According to these characteristics, the information clearly interferes the testing performance of a prediction model. Thus, minimizing the aforementioned dependence is desirable. First, a group of new features need to be designed to describe the background information. The features are called ``domain features''. From the perspective of drift correction, there are two main types of background information: the device label (with which device the sample was collected) and the acquisition time (when the sample was collected). We can actually encode more information such as the place of collection, the operation condition, and so on, which will be useful in other domain adaptation problems.

Formally, if we only consider the instrumental variation, the following one-hot coding scheme can be used. Suppose there are $ n_{dev} $ devices, which result in $ n_{dev} $ different but related domains. The domain feature vector is thus $ \bm{d} \in {\bf{R}}^{n_{dev}} $, where $ d_p=1 $ if the sample is from the $ p $th device and 0 otherwise. If the time-varying drift is also considered, the acquisition time can be further added. If a sample was collected from the $ p $th device at time $ t $, then $ \bm{d} \in {\bf{R}}^{2n_{dev}} $, where
\begin{equation} \label{eq:domainFt}
d_q =
\begin{cases}
1, & q=2p-1, \\
t, & q=2p, \\
0, & \text{otherwise.}
\end{cases}
\end{equation}

According to \Eq{hsicEmp}, the kernel matrix $ K_d $ of the domain features needs to be computed for HSIC. We apply the linear kernel. Suppose $ D =[\bm{d}_1,\ldots,\bm{d}_n] \in {\bf{R}}^{m_d\times n} $, $ m_d $ is the dimension of a domain feature vector. Then
\begin{equation} \label{eq:Kd}
	K_d = D\Trsps D.
\end{equation}

Note that in traditional domain adaptation problems with several discrete domains, the one-hot coding scheme can be applied to construct domain features, because the problems are similar to instrumental variation.

\subsection{Feature Augmentation} \label{subsec:ftAug}

Feature augmentation is used in this paper to learn background-specific subspaces. In \cite{Daume07Frustr}, the author proposed a feature augmentation strategy for domain adaptation by replicating the original features. However, this strategy requires that data lie in discrete domains and cannot deal with time-varying drift. We propose a more general and efficient feature augmentation strategy: concatenating the original features and the domain features, i.e.
\begin{equation} \label{eq:domainFt}
	\hat{\bm{x}} = \left[\begin{matrix}\bm{x} \\ \bm{d}\end{matrix} \right]\in {\bf{R}}^{m+m_d}.
\end{equation}

The role of this strategy can be demonstrated through a linear dimensionality reduction example. Suppose a projection matrix $ W\in {\bf{R}}^{(m+m_d)\times h} $ has been learned for the augmented feature vector. $ h $ is the dimension of the subspace. $ W $ has two parts: $ W=\left[\begin{matrix} W_x \\ W_d \end{matrix} \right], W_x\in {\bf{R}}^{m\times h}, W_d\in {\bf{R}}^{m_d\times h} $. The embedding of $ \hat{\bm{x}} $ can be expressed as $ W\Trsps\hat{\bm{x}} = W_x\Trsps\bm{x} + W_d\Trsps\bm{d} \in {\bf{R}}^h $, which means that a background-specific bias $ (W_d\Trsps\bm{d})_i $ has been added to each dimension $ i $ of the embedding. From another perspective, the feature augmentation strategy maps the samples to an augmented space with higher dimension before projecting them to a subspace. It will be easier to find a projection direction in the augmented space to align the samples well in the subspace.

Take machine olfaction for example, there are situations when the conditional probability $ P(Y|X) $ changes along with the background. For instance, the sensitivity of chemical sensors often decays over time. A signal that indicates low concentration in an earlier time actually suggests high concentration in a later time. In such cases, feature augmentation is important, because it allows samples with similar appearance but different concepts to be treated differently by the background-specific bias. The strategy also helps to align the domains better in each projected dimension. Its effect will be illustrated on several synthetic datasets in Section \ref{subsec:synExp} and further analyzed in the complementary materials.

\subsection{Maximum Independence Domain Adaptation (MIDA)} \label{subsec:mida}

In this section, we introduce the formulation of MIDA in detail. Suppose $ X \in {\bf{R}}^{m\times n} $ is the matrix of $ n $ samples. The training and the test samples are pooled together. More importantly, we do not have to explicitly differentiate which domain a sample is from. The feature vectors have been augmented, but we use the notations $ X$ and $ m $ instead of $ \hat{X}$ and $ m+m_d $ for brevity. A linear or nonlinear mapping function $ \varPhi $ can be used to map $ X $ to a new space. Based on the kernel trick, we need not know the exact form of $ \varPhi $, but the inner product of $ \varPhi(X) $ can be represented by the kernel matrix $ K_x = \varPhi(X)\Trsps\varPhi(X) $. Then, a projection matrix $ \tilde{W} $ is applied to project $ \varPhi(X) $ to a subspace with dimension $ h $, leading to the projected samples $ Z = \tilde{W}\Trsps\varPhi(X) \in {\bf{R}}^{h\times n} $. Similar to other kernel dimensionality reduction algorithms \cite{Schol98Kpca, Schol99Kfda}, the key idea is to express each projection direction as a linear combination of all samples in the space, namely $ \tilde{W} = \varPhi(X)W $. $ W \in {\bf{R}}^{n\times h} $ is the projection matrix to be actually learned. Thus, the projected samples are
\begin{equation} \label{eq:Z}
	Z = W\Trsps\varPhi(X)\Trsps\varPhi(X) = W\Trsps K_x.
\end{equation}
Intuitively, if the projected features are independent of the domain features, then we cannot distinguish the background of a sample by its projected features, suggesting that the inter-domain discrepancy is diminished in the subspace. Therefore, after omitting the scaling factor in \Eq{hsicEmp}, we get the expression to be minimized: $ \text{tr}(K_zHK_dH) = \text{tr}(K_xWW\Trsps K_xHK_dH) $, where $ K_z $ is the kernel matrix of $ Z $.

In domain adaptation, the goal is not only minimizing the difference of distributions, but also preserving important properties of data, such as the variance \cite{Pan11TCA}. It can be achieved by maximizing the trace of the covariance matrix of the project samples. The covariance matrix is
\begin{equation} \label{eq:covZ}
		\text{cov}(Z) = \text{cov}(W\Trsps K_x) = W\Trsps K_xHK_xW,
\end{equation}
where $ H = I-n^{-1}\bm{1}_n\bm{1}_n\Trsps $ is the same as that in \Eq{hsicEmp}. An orthonormal constraint is further added on $ W $. The learning problem then becomes
\begin{equation} \label{eq:midaObj}
	\begin{split}
		\max_W \quad&-\text{tr}(W\Trsps K_xHK_dHK_xW) + \mu\,\text{tr}(W\Trsps K_xHK_xW), \\
		\text{s.t.} \quad&W\Trsps W = I,
	\end{split}
\end{equation}
where $ \mu>0 $ is a trade-off hyper-parameter. Using the Lagrangian multiplier method, we can find that $ W $ is the eigenvectors of $ K_x(-HK_dH+\mu H)K_x $ corresponding to the $ h $ largest eigenvalues. Note that a conventional constraint is requiring $ \tilde{W} $ to be orthonormal as in \cite{Bar11Spca}, which will lead to a generalized eigenvector problem. However, we find that this strategy is inferior to the proposed one in both adaptation accuracy and training speed in practice, so it is not used.

When computing $ K_x $, a proper kernel function needs to be selected. Common kernel functions include linear ($ k(\bm{x},\bm{y})=\bm{x}\Trsps\bm{y} $), polynomial ($ k(\bm{x},\bm{y})=(\sigma\bm{x}\Trsps\bm{y}+1)^d $), Gaussian radial basis function (RBF, $ k(\bm{x},\bm{y})=\exp(\frac{\|\bm{x}-\bm{y}\|^2}{2\sigma^2}) $), and so on. Different kernels indicate different assumptions on the type of dependence in using HSIC \cite{Song12FtSelHsic}. According to \cite{Song12FtSelHsic}, the polynomial and RBF kernels map the original features to a higher or infinite dimensional space, thus are able to detect more types of dependence. However, choosing a suitable kernel width parameter ($ \sigma $) is also important for these more powerful kernels \cite{Song12FtSelHsic}.

The maximum mean discrepancy (MMD) criterion is used in TCA \cite{Pan11TCA} to measure the difference of two distributions. Song et al.\ \cite{Song12FtSelHsic} showed that when HSIC and MMD are both applied to measure the dependence between features and labels in a binary-class classification problem, they are identical up to a constant factor if the label kernel matrix in HSIC is properly designed. However, TCA is feasible only when there are two discrete domains. On the other hand, MIDA can deal with a variety of situations including multiple domains and continuous distributional change. The stationary subspace analysis (SSA) algorithm \cite{Bunau09Ssa} is able to identify temporally stationary components in multivariate time series. However, SSA only ensures that the mean and covariance of the components are stationary, while they may not be suitable for preserving important properties in data. Concept drift adaptation algorithms \cite{Gama14CdaSurvey} are able to correct continuous time-varying drift. However, most of them rely on newly arrived labeled data to update the prediction models, while MIDA works unsupervisedly. 

\subsection{Semi-supervised MIDA (SMIDA)} \label{subsec:smida}

MIDA aligns samples with different backgrounds without considering the label information. However, if the labels of some samples are known, they can be incorporated into the subspace learning process, which may be beneficial to prediction. Therefore, we extend MIDA to semi-supervised MIDA (SMIDA). Since we do not explicitly differentiate the domain labels of the samples, both unlabeled and labeled samples can exist in any domain. Similar to \cite{Song07Cmvu, Pan11TCA, Bar11Spca, Song12FtSelHsic}, HSIC is adopted to maximize the dependence between the projected features and the labels. The biggest advantage of this strategy is that all types of labels can be exploited, such as the discrete labels in classification and the continuous ones in regression.

The label matrix $ Y $ is defined as follows. For $ c $-class classification problems, the one-hot coding scheme can be used, i.e.\ $ Y\in {\bf{R}}^{c\times n}, y_{i,j}=1 $ if $ \bm{x}_i $ is labeled and belongs to the $ j $th class; 0 otherwise. For regression problems, the target values can be centered first. Then, $ Y\in {\bf{R}}^{1\times n}, y_{i} $ equals to the target value of $ \bm{x}_i $ if it is labeled; 0 otherwise. The linear kernel function is chosen for the label kernel matrix, i.e.
\begin{equation} \label{eq:Ky}
	K_y = Y\Trsps Y.
\end{equation}

The objective of SMIDA is
\begin{equation} \label{eq:smidaObj}
	\begin{split}
		\max_W \quad&\text{tr}(W\Trsps K_x(-HK_dH+\mu H+\gamma HK_yH)K_xW), \\
		\text{s.t.} \quad&W\Trsps W = I,
	\end{split}
\end{equation}
where $ \gamma>0 $ is a trade-off hyper-parameter. Its solution is the eigenvectors of $ K_x(-HK_dH+\mu H+\gamma HK_yH)K_x $ corresponding to the $ h $ largest eigenvalues. The outline of MIDA and SMIDA is summarized in Algorithm \ref{algo:smida}. The statements in brackets correspond to those specialized for SMIDA.

\begin{algorithm}[]
	
	\caption{MIDA [or SMIDA]}
	\label{algo:smida}
	
	\begin{algorithmic}[1]
		
		\REQUIRE The matrix of all samples $ X $ and their background information; [the labels of some samples]; the kernel function for $ X $; $ h, \mu, \left[\text{and } \gamma\right] $.
		
		\ENSURE The projected samples $ Z $.
		
		\STATE Construct the domain features according to the background information, e.g.\ Section \ref{subsec:domainFt}.
		
		\STATE Augment the original features with domain features \Eq{domainFt}.
		
		\STATE Compute the kernel matrices $ K_x, K_d \,\Eq{Kd}, \left[\text{and } K_y \,\Eq{Ky}\right]$.
		
		\STATE Obtain $ W $, namely the eigenvectors of $ K_x(-HK_dH+\mu H)K_x $ $ \left[ \text{or } K_x(-HK_dH+\mu H+\gamma HK_yH)K_x \right]$ corresponding to the $ h $ largest eigenvalues.
		
		\STATE $ Z = W\Trsps K_x $.
		
	\end{algorithmic}
\end{algorithm}

Besides variance and label dependence, another useful property of data is the geometry structure, which can be preserved by manifold regularization (MR) \cite{Belkin06MR}. MR can be conveniently incorporated into SMIDA. In our experiments, adding MR generally increases the accuracy slightly with the cost of three more hyper-parameters. Consequently, it is not adopted in this paper.

\section{Experiments} \label{sec:exp}

In this section, we first conduct experiments on some synthetic datasets to verify the effect of the proposed methods. Then, drift correction experiments are performed on two e-nose
datasets and a spectroscopy dataset. To show the universality of the proposed methods, we further evaluate them on a visual object recognition dataset. Comparison is made between them and recent unsupervised domain adaptation algorithms that learn domain-invariant features. 

\subsection{Synthetic Dataset} \label{subsec:synExp}

In \Fig{synExp_twoDomain}, TCA \cite{Pan11TCA} and MIDA are compared on a 2D dataset with two discrete domains. The domain labels were used to construct the domain features in MIDA according to the one-hot coding scheme introduced in Section \ref{subsec:domainFt}. The similar definition was used in synthetic datasets 3 and 4. For both methods, the linear kernel was used on the original features and the hyper-parameter $ \mu $ was set to 1. In order to quantitatively assess the effect of domain adaptation, logistic regression models were trained on the labeled source data and tested on the target data. The accuracies are displayed in the caption, showing that the order of performance is MIDA $ > $ TCA $ > $ original feature. TCA aligns the two domains only on the first projected dimension. However, the two classes have large overlap on that dimension, because the direction for alignment is different from that for discrimination. Incorporating the label information of the source domain (SSTCA) did no help. On the contrary, MIDA can align the two domains well in both projected dimensions, in which the domain-specific bias on the second dimension brought by feature augmentation played a key role. A 3D explanation is included in the supplementary materials. Thus, good accuracy can be obtained by using the two dimensions for classification.

\def \figWidth {5.9in}
\begin{figure*}[]
	\centering
	\includegraphics[trim=1in 0in 1in 0.1in,clip,width=\figWidth]{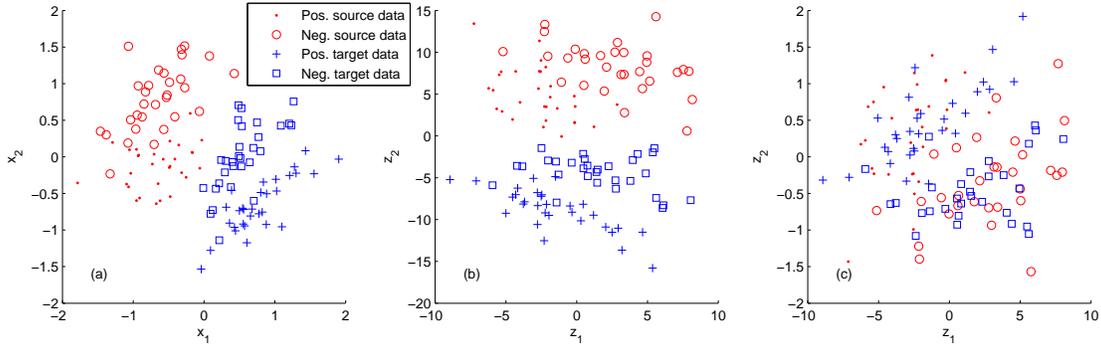} 
	\caption{Comparison of TCA and MIDA in a 2D synthetic dataset. Plots (a)-(c) show data in the original space and projected spaces of TCA and MIDA, respectively. The classification accuracies are 53\%, 70\% (only using the first projected dimension $ z_1 $), and 88\%.}
	\label{fig:synExp_twoDomain} 
\end{figure*}

\begin{figure*}[]
	\centering
	\includegraphics[trim=1in 0in 1in 0.1in,clip,width=\figWidth]{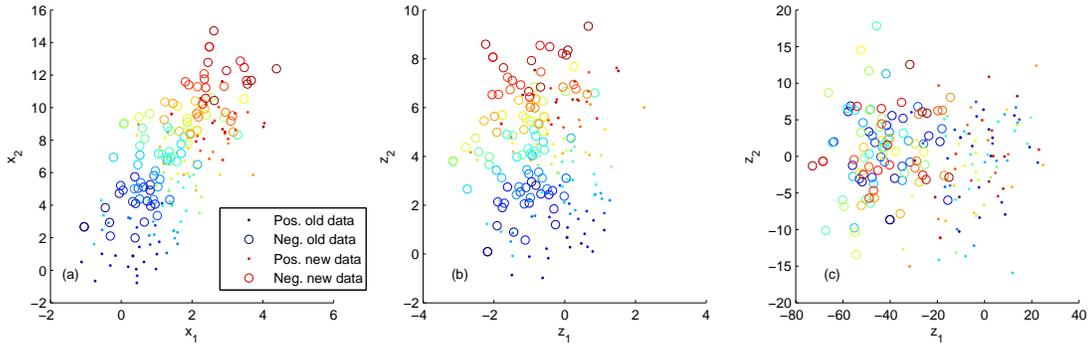}
	\caption{Comparison of SSA and MIDA in a 2D synthetic dataset. Plots (a)-(c) show data in the original space, projected spaces of SSA and MIDA, respectively. The chronological order of a sample is indicated by color. The classification accuracies are 55\%, 74\% (only using the first projected dimension $ z_1 $), and 90\%.}
	\label{fig:synExp_time} 
\end{figure*}

In \Fig{synExp_time}, SSA \cite{Bunau09Ssa} and MIDA are compared on a 2D dataset with continuous distributional change, which resembles time-varying drift in machine olfaction. Samples in both classes drift to the upper right. The chronological order of the samples was used to construct the domain features in MIDA, i.e. $ d=1 $ for the first sample, $ d=2 $ for the second sample, etc. The parameter setting of MIDA was the same with that in \Fig{synExp_twoDomain}, whereas the number of stationary components in SSA was set to 1. The classification accuracies were obtained by training a logistic regression model on the first halves of the data in both classes, and testing them on the last halves. SSA succeeds in finding a direction ($ z_1 $) that is free from time-varying drift. However, the two classes cannot be well separated in that direction. In plot (c), the randomly scattered colors suggest that the time-varying drift is totally removed in the subspace. MIDA first mapped the 2D data into a 3D space with the third dimension being time, then projected them to a 2D plane orthogonal to the direction of drift in the 3D space.

\begin{figure*}[]
	\centering
	\includegraphics[trim=1in 0in 1in 0.1in,clip,width=\figWidth]{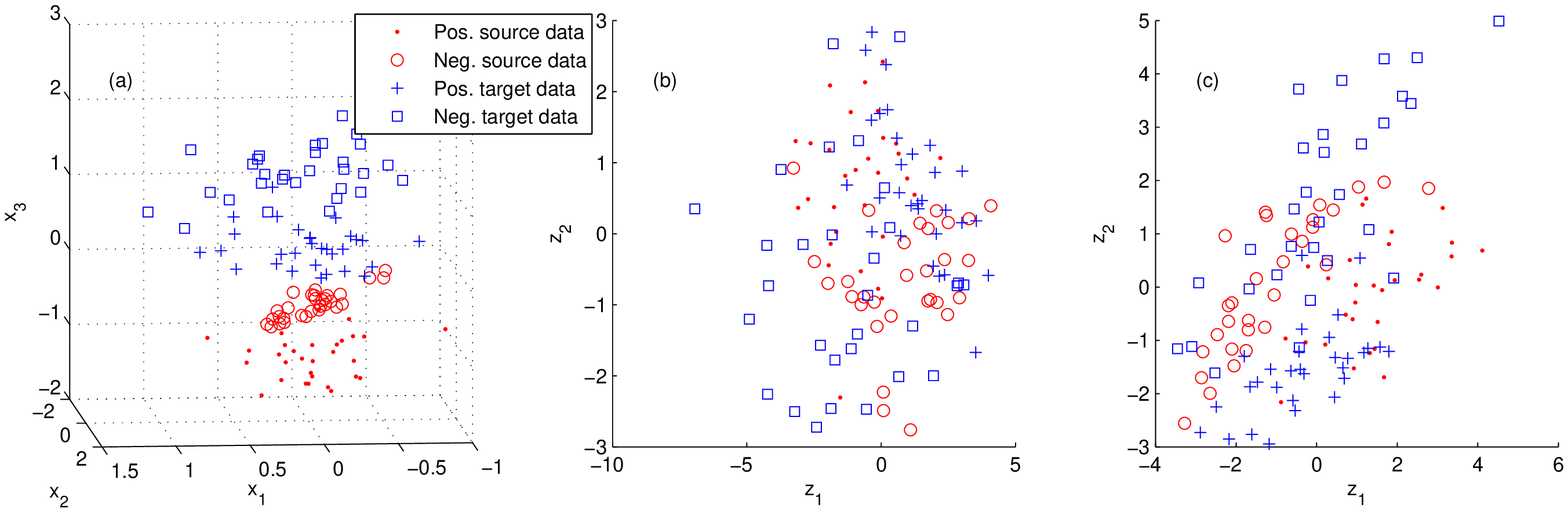}
	\caption{Comparison of MIDA and SMIDA in a 3D synthetic dataset. Plots (a)-(c) show data in the original space and projected spaces of MIDA and SMIDA, respectively. The classification accuracies are 50\%, 55\%, and 82\%.}
	\label{fig:synExp_label} 
\end{figure*}

No label information was used in the last two experiments. If keeping the label dependence in the subspace is a priority, SMIDA can be adopted instead of MIDA. In the 3D synthetic dataset in \Fig{synExp_label}, the best direction ($ x_3 $) to align the two domains also mixes the two classes, which results in the output of MIDA in plot (b). The labels in the source domain were used when learning the subspace. From plot (c), we can observe that the classes are separated. In fact, class separation can still be found in the third dimension of the space learned by MIDA. However, for the purpose of dimensionality reduction, we generally hope to keep the important information in the first few dimensions.

\begin{figure*}[]
	\centering
	\includegraphics[trim=1in 0in 1in 0.1in,clip,width=\figWidth]{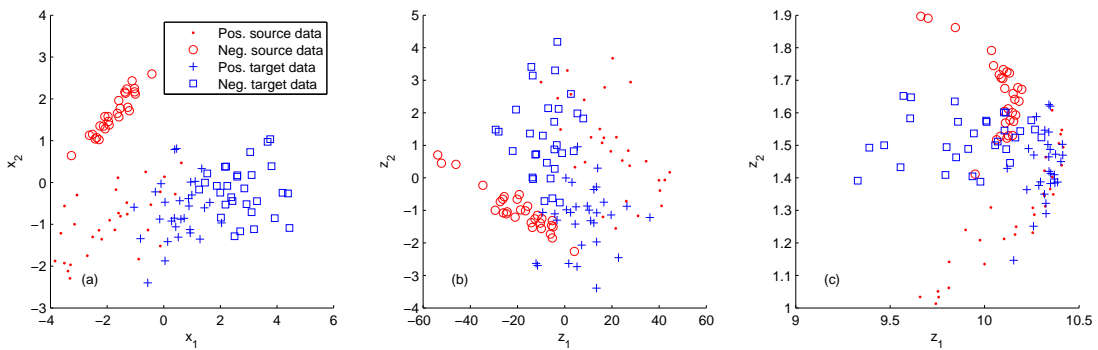}
	\caption{Comparison of different kernels in a 2D synthetic dataset. Plots (a)-(c) show data in the original space and projected spaces of MIDA with linear and RBF kernels, respectively. The classification accuracies are 50\%, 57\%, and 87\%.}
	\label{fig:synExp_nonlin} 
\end{figure*}

Nonlinear kernels are often applied in machine learning algorithms when data is not linearly separable. Besides, they are also useful in domain adaptation when domains are not linearly ``alignable'', as shown in \Fig{synExp_nonlin}. In plot (a), the inter-domain changes in distributions are different for the two classes. Hence, it is difficult to find a linear projection direction to align the two domains, even with the domain-specific biases of MIDA. Actually, domain-specific rotation matrices are needed. Since the target labels are not available, the rotation matrices cannot be obtained accurately. However, a nonlinear kernel can be used to map the original features to a space with higher dimensions, in which the domains may be linearly alignable. We applied an RBF kernel with width $ \sigma=10 $. Although the domains are not perfectly aligned in plot (c), the classification model trained in the source domain can be better adapted to the target domain. A comparison on different kernel and kernel parameters on two synthetic datasets is included in the supplementary materials.

\subsection{Gas Sensor Array Drift Dataset} \label{subsec:expUci}

The gas sensor array drift dataset\footnote{{ http://archive.ics.uci.edu/ml/datasets/\\Gas+Sensor+Array+Drift+Dataset+at+Different+Concentrations}} collected by Vergara et al. \cite{Vergara12Ensemble} is dedicated to research in drift correction. A total of 13910 samples were collected by an e-nose with 16 gas sensors over a course of 36 months. There are six different kinds of gases at different concentrations. They were split into 10 batches by the authors according to their acquisition time. Table A in the supplementary material details the dataset. We aim to classify the type of gases, despite their concentrations. 

\begin{figure}[]
	\centering
	\includegraphics[trim=20 0mm 30 0,clip,width=3.5in]{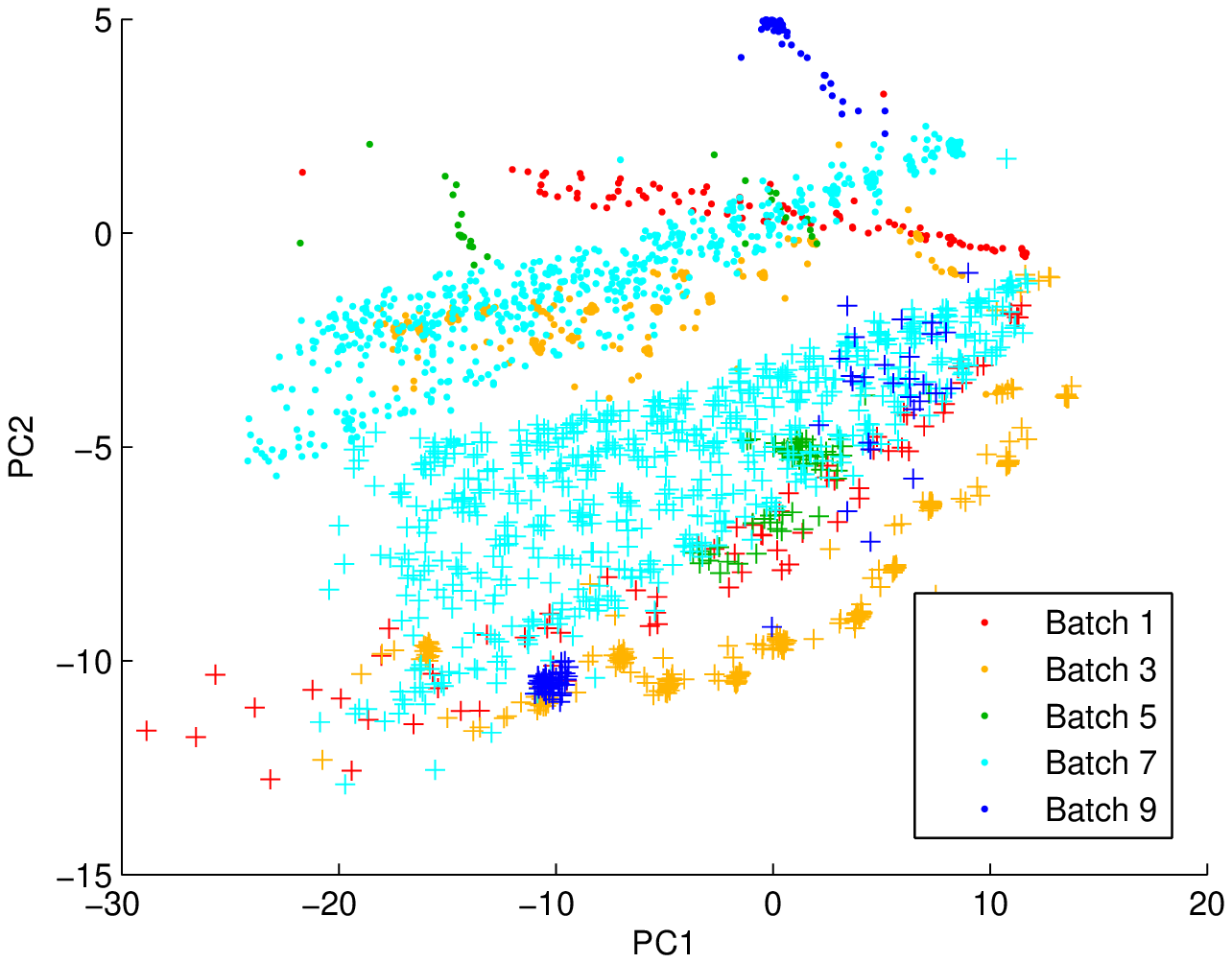} 
	\caption{Scatter of ethanol (dots) and acetone (plus signs) samples in batches 1,3,5,7,9 in the gas sensor array drift dataset. Samples are projected to a 2D subspace using PCA. Different colors indicate different batches.}
	\label{fig:uciDrift}
\end{figure}

Similar to \cite{Vergara12Ensemble, Liu14Semi}, we took the samples in batch 1 as labeled training samples, whereas those in batches 2--10 are unlabeled test ones. This evaluation strategy resembles the situation in real-world applications. In the dataset, each sample is represented by 128 features extracted from the sensors' response curves \cite{Vergara12Ensemble}. Each feature was first normalized to zero mean and unit variance within each batch. The time-varying drift of the preprocessed features across batches can be visually inspected in \Fig{uciDrift}. It is obvious that samples in different batches have different distributions. Next, the labeled samples in batch 1 were adopted as the source domain and the unlabeled ones in batch $ b $ ($ b = 2, \ldots, 10 $) as the target domain. The proposed algorithms together with several recent ones were used to learn domain-invariant features based on these samples. Then, a logistic regression model was trained on the source domain and tested on each target one. For multi-class classification, the one-vs-all strategy was utilized.

As displayed in \Table{uciAcc}, the compared methods include kernel PCA (KPCA), transfer component analysis (TCA), semi-supervised TCA (SSTCA) \cite{Pan11TCA}, subspace alignment (SA) \cite{Fern13Sa}, geodesic flow kernel (GFK) \cite{Gong12Gfk}, manifold regularization with combination GFK (ML-comGFK) \cite{Liu14Semi}, information-theoretical learning (ITL) \cite{Shi12Itl}, structural correspondence learning (SCL) \cite{Blit07SCLMI}, and marginalized stacked denoising autoencoder (mSDA) \cite{Chen12mSDA}. For all methods, the hyper-parameters were tuned for the best accuracy. In KPCA, TCA, SSTCA, and the proposed MIDA and SMIDA, the polynomial kernel with degree 2 was used. KPCA learned a subspace based on the union of source and target data. In TCA, SSTCA, MIDA, and SMIDA, eigenvalue decomposition needs to be done on kernel matrices. In order to reduce the computational burden, we randomly chose at most $ n_t $ samples in each target domain when using these methods, with $ n_t $ being twice the number of the samples in the source domain. GFK used PCA to generate the subspaces in both source and target domains. The subspace dimension of GFK was determined according to the subspace disagreement measure in \cite{Gong12Gfk}. The results of ML-comGFK are copied from \cite{Liu14Semi}. In SCL, the pivot features were binarized before training pivot predictors using logistic regression.

We also compared several variants of our methods. In \Table{uciAcc}, the notation ``(discrete)'' means that two discrete domains (source and target) were used in MIDA and SMIDA, which is similar to other compared methods. The domain feature vector of a sample was thus $ [1,0]\Trsps $ if it was from the source domain and $ [0,1]\Trsps $ if it was from the target. However, this strategy cannot make use of the samples in intermediate batches. An intuitive assumption is that the distributions of adjacent batches should be similar. When adapting the information from batch 1 to $ b $, taking samples from batches 2 to $ b-1 $ into consideration may improve the generalization ability of the learned subspace. Concretely, $ n_t $ samples were randomly selected from batches 2 to $ b $ instead of batch $ b $ alone. For each sample, the domain feature was defined as its batch index, which can be viewed as a proxy of its acquisition time. MIDA and SMIDA then maximized the independence between the learned subspace and the batch indices. The results are labeled as ``(continuous)'' in \Table{uciAcc}. Besides, the accuracies of continuous SMIDA without feature augmentation (no aug.) are also shown.

\begin{table*}[] 
	\renewcommand{\tabcolsep}{0.15cm}
	\renewcommand{\arraystretch}{1.5}
	\renewcommand{\doublerulesep}{0pt}
	\caption{Classification accuracy (\%) on the gas sensor array drift dataset. Bold values indicate the best results.}
	\label{tbl:uciAcc}
	\centering
	\begin{tabular}{lllllllllll}
		\hline \hline
		& Batch	2	& 3	& 4	& 5	& 6	& 7	& 8	& 9	& 10 & Average	\\
		\hline
		Original feature& 80.47	& 79.26	& 69.57	& 77.16	& 77.39	& 64.21	& 52.04	& 47.87	& 48.78	& 66.30 \\
		KPCA			& 75.88	& 69.04	& 49.07	& 57.87	& 62.65	& 52.26	& 37.07	& 47.66	& 49.97	& 55.72 \\
		TCA \cite{Pan11TCA}		& 82.96	& 81.97	& 65.22	& 76.14	& 89.09	& 58.98	& 49.32	& 66.17	& 49.50	& 68.82 \\
		SSTCA \cite{Pan11TCA}	& \bf 84.57	& 80.90	& \bf 80.12	& 75.63	& 87.26	& 66.37	& 54.76	& 61.28	& 54.44	& 71.70 \\
		SA \cite{Fern13Sa}		& 80.79	& 80.01	& 71.43	& 75.63	& 78.35	& 64.68	& 52.04	& 48.51	& 49.58	& 66.78 \\
		GFK \cite{Gong12Gfk}	& 77.41	& 80.26	& 71.43	& 76.14	& 77.65	& 64.99	& 36.39	& 47.45	& 48.72	& 64.49 \\
		ML-comGFK \cite{Liu14Semi}	& 80.25	& 74.99	& 78.79	& 67.41	& 77.82	& \bf 71.68	& 49.96	& 50.79	& 53.79	& 67.28 \\
		ITL \cite{Shi12Itl}		& 76.85	& 79.45	& 59.63	& \bf 96.45	& 78.00	& 60.95	& 49.32	& \bf 77.02	& 48.58	& 69.58 \\
		SCL \cite{Blit07SCLMI}	& 77.57	& 82.03	& 68.32	& 82.74	& 77.22	& 65.18	& 53.74	& 48.51	& 48.08	& 67.04 \\
		mSDA \cite{Chen12mSDA}	& 73.87	& 79.19	& 65.84	& 80.20	& 76.39	& 65.90	& 51.70	& 48.51	& 48.92	& 65.61 \\
		\hline
		MIDA (discrete)	& 81.03	& 85.62	& 60.25	& 75.63	& 87.61	& 62.44	& 48.30	& 67.87	& 48.36	& 68.57 \\
		SMIDA (discrete)& 80.47	& \bf 87.07	& 65.22	& 75.63	& 90.04	& 59.20	& 50.00	& 62.77	& 44.81	& 68.36 \\
		MIDA (continuous)		& 84.32	& 81.59	& 68.32	& 75.63	& 91.74	& 63.13	& 78.91	& 62.34	& 45.14	& 72.35 \\
		SMIDA (no aug.)	& 82.23	& 83.17	& 67.70	& 75.13	& 85.22	& 61.67	& 51.02	& 61.49	& \bf 54.61	& 69.14 \\
		SMIDA (continuous)		& 83.68	& 82.28	& 73.91	& 75.63	& \bf 93.00	& 63.49	& \bf 79.25	& 62.34	& 45.50	& \bf 73.23 \\
		\hline \hline
	\end{tabular}
\end{table*}

\begin{figure}[]
	\centering
	\includegraphics[trim=20 0mm 30 0,clip,width=3in]{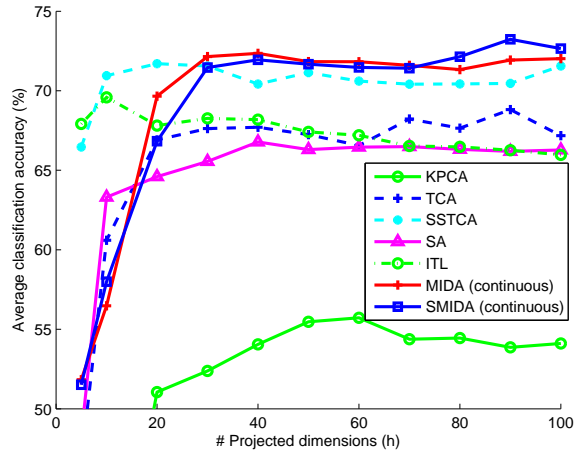} 
	\caption{Performance comparison on the gas sensor array drift dataset with respect to the subspace dimension $ h $.}
	\label{fig:uciAccVsDim}
\end{figure}

From \Table{uciAcc}, we can find that as the batch index increases, the accuracies of all methods generally degrade, which confirms the influence of the time-varying drift. Continuous SMIDA achieves the best average domain adaptation accuracy. The continuous versions of MIDA and SMIDA outperform the discrete versions, proving that the proposed methods can effectively exploit the chronological information of the samples. They also surpass ML-comGFK which uses the samples in intermediate batches to build connections between the source and the target batches. Feature augmentation is important in this dataset, since removing it in continuous SMIDA causes a drop of four percentage points in average accuracy. In \Fig{uciAccVsDim}, the average classification accuracies with varying subspace dimension are shown. MIDA and SMIDA are better than other methods when more than 30 features are extracted.

\subsection{Breath Analysis Dataset} \label{subsec:expBreath}

As a noninvasive approach, disease screening and monitoring with e-noses is attracting more and more attention \cite{Rock08Review, Yan14System}. The concentration of some biomarkers in breath has been proved to be related to certain diseases, which makes it possible to analyze a person's health state with an e-nose conveniently. For example, the concentration of acetone in diabetics' breath is often higher than that in healthy people \cite{Yan14System}. However, the instrumental variation and time-varying drift of e-noses hinder the popularization of this technology in real-world applications. Unsupervised domain adaptation algorithms can be applied to solve this problem.

We have collected a breath analysis dataset in years 2014--2015 using two e-noses of the same model \cite{Yan14System}. In this paper, samples of five diseases were selected for experiments, including diabetes, chronical kidney disease (CKD), cardiopathy, lung cancer, and breast cancer. They have been proved to be related to certain breath biomarkers. We performed five binary-class classification tasks to distinguish samples with one disease from the healthy samples. Each sample was represented by the steady state responses of nine gas sensors in the e-nose. When a gas sensor is used to sense a gas sample, its response will reach a steady state in a few minutes. The steady state response has a close relationship with the concentration of the measured gas. Therefore, the 9D feature vector contains most information needed for disease screening.

\begin{figure}[]
	\centering
	\includegraphics[trim=20 0mm 30 0,clip,width=3in]{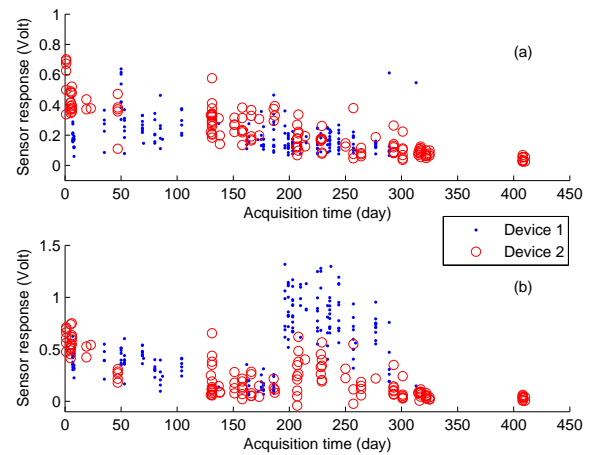} 
	\caption{Illustration of the instrumental variation and time-varying drift in the breath analysis dataset. Plots (a) and (b) show the steady state responses of the CKD samples of sensors 2 and 7, respectively.}
	\label{fig:breathDrift}
\end{figure}

To show the instrumental variation and time-varying drift in the dataset, we draw the steady state responses of two sensors of the CKD samples in \Fig{breathDrift}. Each data point indicates a breath sample. In plot (a), the sensitivity of the sensor in both devices gradually decayed as time elapsed. In plot (b), the aging effect was so significant that we had to replace the sensors in the two devices with new ones on about day 200. In this case, a signal at 0.3 V will suggest low concentration on day 0 but high concentration on day 150.  In addition, the responses in different devices are different (e.g.\ plot (b), after day 200).

The numbers of samples in the six classes (healthy and the five diseases mentioned above) are 125, 431, 340, 97, 156, and 215, respectively. We chose the first 50 samples collected with device 1 in each class as labeled training samples. Among the other samples, 10 samples were randomly selected in each class for validation, the rest for testing. The hyper-parameters were tuned on the validation sets. Logistic regression was adopted as the classifier, with F-score as the accuracy criterion. Results are compared in \Table{breathAcc}.

In KPCA, TCA, SSTCA, MIDA, and SMIDA, the RBF kernel was used. Because methods other than stationary subspace analysis (SSA) \cite{Bunau09Ssa}, MIDA, and SMIDA are not capable of handling the chronological information, we simply regarded each device as a discrete domain and learned device-invariant features with them. The same strategy was used in discrete MIDA and SMIDA. In continuous MIDA and SMIDA, the domain features were defined according to \Eq{domainFt}, where $ t $ was the exact acquisition time converted to years and the number of devices $ n_{dev}=2 $. SSA naturally considers the chronological information by treating the sample stream as a multivariate time series and identifying temporally stationary components. However, SSA cannot deal with time series with multiple sources, such as the multi-device case in this dataset. Thus, the samples were arranged in chronological order despite their device labels.

\begin{table}[]
	\renewcommand{\tabcolsep}{0.15cm}
	\renewcommand{\arraystretch}{1.5}
	\renewcommand{\doublerulesep}{0pt}
	\caption{Classification accuracy (\%) on the breath analysis dataset. Bold values indicate the best results.}
	\label{tbl:breathAcc}
	\centering
	\begin{tabular}{lllllll}
		\hline \hline
								& Task 1& 2		& 3		& 4		& 5		& Average	\\
		\hline
		Original feature		& 34.34	& 63.67	& 73.71	& 43.17	& 42.93	& 51.57 \\
		KPCA					& 58.05	& 72.58	& 84.78	& 44.95	& 42.60	& 60.59 \\
		TCA \cite{Pan11TCA}		& 67.19	& 68.31	& 59.93	& 67.08	& \bf 68.17	& 66.14 \\
		SSTCA  \cite{Pan11TCA}	& 67.01	& 68.06	& 74.14	& 68.31	& 67.36	& 68.97 \\
		SA \cite{Fern13Sa}		& 29.95	& 72.42	& 72.74	& 42.19	& 44.54	& 52.37 \\
		GFK \cite{Gong12Gfk}	& 41.49	& 68.50	& 58.96	& 75.63	& 70.16	& 62.95 \\
		ITL \cite{Shi12Itl}		& 68.59	& 66.53	& 74.75	& 66.67	& 68.03	& 68.91 \\
		SSA \cite{Bunau09Ssa}	& 49.77	& 72.10	& 33.49	& 52.64	& 55.38	& 52.68 \\
		SCL \cite{Blit07SCLMI}	& 32.52 & 61.16 & 75.43 & 35.35 & 51.86 & 51.26 \\
		mSDA \cite{Chen12mSDA}	& 36.86	& 69.51	& 76.69	& 35.51	& 50.49	& 53.81 \\
		
		\hline
		MIDA (discrete)			& 62.17	& 71.74	& 84.21	& 67.05	& 67.06	& 70.45 \\
		SMIDA (discrete)		& 80.16	& \bf 84.18	& 88.47	& 68.45	& 52.41	& 74.73 \\
		MIDA (continuous)		& 68.30	& 67.54	& 74.01	& 73.04	& 69.63	& 70.50 \\
		SMIDA (no aug.)			& 82.80	& 72.57	& 72.61	& \bf 80.33	& 70.05	& 75.67 \\
		SMIDA (continuous)		& \bf 85.29	& 80.18	& \bf 91.67	& 74.28	& 66.55	& \bf 79.59 \\
		\hline \hline
	\end{tabular}
\end{table}

From \Table{breathAcc}, we can find that the improvement made by SSA is little, possibly because the stationary criterion is not suitable for preserving important properties in data. For example, the noise in data can also be stationary \cite{Pan11TCA}. MIDA and SMIDA achieved obviously better results than other methods. They can address both instrumental variation and time-varying drift. With the background-specific bias brought by feature augmentation, they can compensate for the change in conditional probability in this dataset. SMIDA is better than MIDA because the label information of the first 50 samples in each class was better kept.

\subsection{Corn Dataset} \label{subsec:expCorn}

Similar to e-noses, data collected with spectrometers are one-dimensional signals indicating the concentration of the analytes. Instrumental variation is also a problem for them \cite{Feud02TransRev}. In this section, we test our methods on the corn dataset\footnote{http://www.eigenvector.com/data/Corn/}. It is a spectroscopy dataset collected with three near-infrared spectrometers designated as m5, mp5, and mp6. The moisture, oil, protein, and starch contents of 80 corn samples were measured by each device, with ranges of the measured values as [9.377, 10.993], [3.088, 3.832], [7.654, 9.711], and [62.826, 66.472], respectively. Each sample is represented by a spectrum with 700 features. This dataset resembles traditional domain adaptation datasets because there is no time-varying drift. Three discrete domains can be defined based on the three devices. We adopt m5 as the source domain, mp5 and mp6 as the target ones. In each domain, samples $ 4, 8, \ldots, 76, 80 $ were assigned as the test set, the rest as the training set. For hyper-parameter tuning, we applied a three-fold cross-validation on the training sets of the three domains. After the best hyper-parameters were determined for each algorithm, a regression model was trained on the training set from the source domain and applied on the test set from the target domains. The regression algorithm was ridge regression with the L2 regularization parameter $ \lambda=1 $.

\begin{figure}[]
	\centering
	\includegraphics[trim=10 0mm 20 20,clip,width=3in]{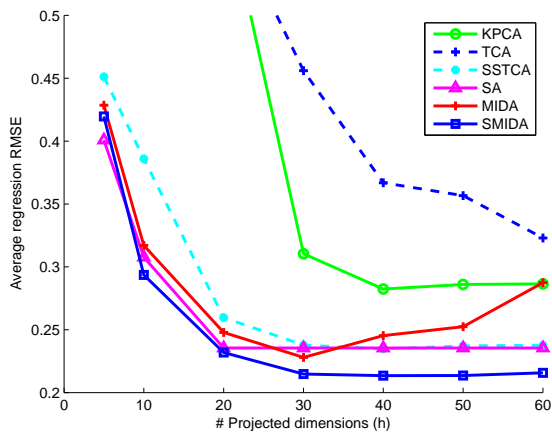} 
	\caption{Performance comparison on the corn dataset with respect to the subspace dimension $ h $.}
	\label{fig:cornAccVsDim}
\end{figure}

\Table{cornRmse} displays the root mean square error (RMSE) of the four prediction tasks and their average on the two target domains. We also plot the overall average RMSE of the two domains with respect to the subspace dimension $ h $ in \Fig{cornAccVsDim}. ITL was not investigated because it is only applicable in classification problems. In KPCA, TCA, SSTCA, MIDA, and SMIDA, the RBF kernel was used. For the semi-supervised methods SSTCA and SMIDA, the target values were normalized to zero mean and unit variance before subspace learning. The domain features were defined according to the device indices using the one-hot coding scheme. We can find that when no domain adaptation was done, the prediction error is large. All domain adaptation algorithms managed to significantly reduce the error. KPCA also has good performance, which is probably because the source and the target domains have similar principal directions, which also contain the most discriminative information. Therefore, source regression models can fit the target samples well. In this dataset, different domains have identical data composition. As a result, corresponding data can be aligned by subspaces alignment, which explains the small error of SA. However, this condition may not hold in other datasets. 

MIDA and SMIDA obtained the lowest average errors in both target domains. Aiming at exploring the prediction accuracy when there is no instrument variation, we further trained regression models on the training set of the two target domains and tested on the same domain. The results are listed as ``train on target'' in \Table{cornRmse}. It can be found that SMIDA outperforms these results. This could be attributed to three reasons: (1) The inter-domain discrepancy in this dataset is relatively easy to correct; (2) The use of RBF kernel in SMIDA improves the accuracy; (3) SMIDA learned the subspace on the basis of both training and test samples. Although the test samples were unlabeled, they can provide some information about the distribution of the samples to make the learned subspace generalize better, which can be viewed as the merit of semi-supervised learning. To testify this assumption, we conducted another experiment with multiple target domains. The training samples from the source domain and the test ones from both target domains were leveraged together for subspace learning in MIDA and SMIDA. The average RMSE for the two target domains are 0.209 and 0.217 for MIDA, and 0.208 and 0.218 for SMIDA. Compared with the results in \Table{cornRmse} with single target domain, the results have been further improved, showing that incorporating more unlabeled samples from target domains can be beneficial.

\begin{table*}[]
	\renewcommand{\tabcolsep}{0.12cm}
	\renewcommand{\arraystretch}{1.5}
	\renewcommand{\doublerulesep}{0pt}
	\caption{Regression RMSE on the corn dataset. Bold values indicate the best results.}
	\label{tbl:cornRmse}
	\centering
	\begin{tabular}{llllllllllll}
		\hline\hline
		& \multicolumn{5}{l}{Mp5 as target domain}	& & \multicolumn{5}{l}{Mp6 as target domain}	\\
		\cline{2-6} \cline{8-12}
								& Moisture& Oil	& Protein& Starch& Average	&& Moisture	& Oil	& Protein	& Starch & Average \\
		\hline
		Original feature		& 1.327	& 0.107	& 1.155	& 2.651	& 1.310	&& 1.433	& 0.101	& 1.413	& 2.776	& 1.431 \\
		KPCA					& 0.477	& 0.165	& 0.215	& \bf 0.315	& 0.293	&& 0.396	& 0.164	& 0.238	& \bf 0.290	& 0.272 \\
		TCA \cite{Pan11TCA}		& 0.539	& 0.322	& 0.217	& 0.402	& 0.370	&& 0.398	& 0.145	& 0.259	& 0.572	& 0.343 \\
		SSTCA  \cite{Pan11TCA}	& 0.343	& 0.093	& \bf 0.140	& 0.366	& 0.235	&& 0.367	& 0.088	& 0.186	& 0.318	& 0.240 \\
		SA \cite{Fern13Sa}		& 0.302	& 0.094	& 0.186	& 0.351	& 0.233	&& 0.324	& 0.079	& 0.158	& 0.390	& 0.238 \\
		GFK \cite{Gong12Gfk}	& 0.267	& 0.197	& 0.342	& 0.621	& 0.357	&& \bf 0.263	& 0.189	& 0.264	& 0.485	& 0.301 \\
		SCL \cite{Blit07SCLMI}	& 0.283	& 0.115	& 0.249	& 0.619	& 0.316 && 0.311 	& 0.108 	& 0.257 	& 0.683 	& 0.340 \\
		mSDA \cite{Chen12mSDA}	& \bf 0.264 & 0.107 & 0.211 & 0.446 & 0.257 && 0.285 	& 0.097 	& 0.198 	& 0.471 	& 0.263 \\
		
		MIDA					& 0.317	& 0.078	& 0.141	& 0.378	& 0.228	&& 0.317	& 0.084	& 0.158	& 0.352	& 0.228 \\
		SMIDA					& 0.287	& \bf 0.072	& 0.143	& 0.339	& \bf 0.210	&& 0.316	& \bf 0.073	& \bf 0.152	& 0.325	& \bf 0.217 \\
		\hline
		Train on target			& 0.176	& 0.094	& 0.201	& 0.388	& 0.215	&& 0.182	& 0.108	& 0.206	& 0.414	& 0.228 \\
		\hline \hline
	\end{tabular}
\end{table*}

\subsection{Visual Object Recognition Dataset} \label{subsec:expVis}

\def \rar {$ \,\rightarrow\, $}

In \cite{Gong12Gfk}, Gong et al.\ evaluated domain adaptation algorithms on four visual object recognition datasets, namely Amazon (A), Caltech-256 (C), DSLR (D), and Webcam (W). Ten common classes were selected from them, with 8 to 151 samples per class per domain, and 2533 images in total. Each image was encoded with an 800-bin histogram using SURF features. The normalized histograms were z-scored to have zero mean and unit variance in each dimension. Following the experimental setting provided in the sample code from the authors of \cite{Gong12Gfk}, experiments were conducted in 20 random trials for each pair of domains. For each unsupervised trail, 20 (for A, C, W) or 8 (for D) labeled samples per class were randomly chosen from the source domain as the training set (other samples were used unsupervisedly for domain adaptation), while all unlabeled samples in the target domain made up the test set. In semi-supervised trails, three labeled samples per class in the target domain were also assumed to be labeled. Averaged accuracies on each pair of domains as well as standard errors are listed in Tables \ref{tbl:visAccUnsup} and \ref{tbl:visAccSemisup}.

\begin{table*}[]
	\renewcommand{\tabcolsep}{0.1cm}
	\renewcommand{\arraystretch}{1.5}
	\renewcommand{\doublerulesep}{0pt}
	\caption{Unsupervised  domain adaptation accuracy (\%) on the visual object recognition dataset. Bold values indicate the best results. X\rar Y means that X is the source domain and Y is the target one.}
	\label{tbl:visAccUnsup}
	\centering
	\scriptsize 
	\begin{tabular}{l lllll lllll lll}
		\hline \hline
			& C\rar A	& D\rar A	& W\rar A	& A\rar C	& D\rar C	& W\rar C	& A\rar D	& C\rar D	& W\rar D	& A\rar W	& C\rar W	& D\rar W & Average \\	
		\hline
Ori. ft.	& 43.2$\pm$2.2	& 34.9$\pm$1.1	& 36.8$\pm$0.6	& 38.5$\pm$1.6	& 31.7$\pm$1.2	& 32.7$\pm$0.9	& 37.3$\pm$3.1	& 40.5$\pm$3.6	& 80.6$\pm$2.0	& 37.5$\pm$2.9	& 37.1$\pm$3.6	& 76.7$\pm$2.0	& 43.97 \\
KPCA	& 27.4$\pm$2.0	& 27.0$\pm$1.6	& 27.6$\pm$1.2	& 25.8$\pm$1.3	& 22.0$\pm$2.4	& 23.7$\pm$1.6	& 29.2$\pm$2.9	& 27.6$\pm$2.7	& 55.1$\pm$1.8	& 29.0$\pm$2.7	& 27.1$\pm$3.2	& 50.3$\pm$2.6	& 30.97 \\
TCA \cite{Pan11TCA}	& 49.8$\pm$2.7	& 38.6$\pm$1.4	& 39.2$\pm$1.0	& 42.8$\pm$2.1	& 35.3$\pm$1.5	& 35.7$\pm$0.8	& \textbf{42.8}$\pm$3.2	& 45.9$\pm$3.8	& 83.9$\pm$1.7	& 41.7$\pm$3.3	& 42.8$\pm$5.4	& 81.5$\pm$2.1	& 48.35 \\
SSTCA \cite{Pan11TCA}	& 50.5$\pm$2.8	& 39.3$\pm$1.6	& 40.5$\pm$0.7	& 42.4$\pm$1.8	& 36.1$\pm$1.5	& 35.9$\pm$0.8	& 42.8$\pm$3.1	& 46.6$\pm$3.5	& 80.6$\pm$2.3	& 42.5$\pm$2.7	& 42.8$\pm$4.7	& 81.8$\pm$1.9	& {48.48} \\
ITL \cite{Shi12Itl}	& 41.2$\pm$3.0	& 35.7$\pm$2.0	& 38.4$\pm$1.1	& 34.8$\pm$1.8	& 28.7$\pm$1.5	& 31.4$\pm$1.6	& 34.5$\pm$3.0	& 32.3$\pm$3.8	& 67.0$\pm$2.7	& 31.8$\pm$3.6	& 36.4$\pm$4.2	& 71.1$\pm$3.2	& 40.27 \\
SA \cite{Fern13Sa}	& 48.4$\pm$2.9	& 36.3$\pm$2.6	& 37.3$\pm$1.7	& 38.5$\pm$2.1	& 33.4$\pm$1.9	& 35.1$\pm$0.9	& 35.0$\pm$3.6	& 39.7$\pm$5.0	& 63.2$\pm$2.8	& 36.7$\pm$4.4	& 41.3$\pm$5.5	& 70.3$\pm$2.5	& 42.93 \\
GFK \cite{Gong14Gfk}	& 40.4$\pm$0.7	& 36.2$\pm$0.4	& 35.5$\pm$0.7	& 37.9$\pm$0.4	& 32.7$\pm$0.4	& 29.3$\pm$0.4	& 35.1$\pm$0.8	& 41.1$\pm$1.3	& 71.2$\pm$0.9	& 35.7$\pm$0.9	& 35.8$\pm$1.0	& 79.1$\pm$0.7	& 42.50 \\
LTSL \cite{Shao14Ltsl}	& 50.4$\pm$0.4	& \textbf{40.2}$\pm$0.6	& \textbf{44.1}$\pm$0.3	& 38.6$\pm$0.3	& 35.3$\pm$0.3	& 37.4$\pm$0.2	& 38.3$\pm$1.1	& \textbf{53.7}$\pm$0.9	& 79.8$\pm$0.4	& 38.8$\pm$1.3	& \textbf{47.0}$\pm$1.0	& 72.8$\pm$0.7	& 48.03 \\
DASC \cite{Cui14Flow}	& 39.1$\pm$0.3	& 39.3$\pm$0.8	& 37.7$\pm$0.7	& \textbf{49.8}$\pm$0.4	& \textbf{48.5}$\pm$0.8	& \textbf{45.4}$\pm$0.9	& 36.5$\pm$0.3	& 35.6$\pm$0.3	& \textbf{88.3}$\pm$0.4	& 36.3$\pm$0.4	& 33.3$\pm$0.3	& 79.8$\pm$0.9	& 47.47 \\
SCL \cite{Blit07SCLMI}	& 43.5$\pm$2.2	& 34.7$\pm$1.3	& 36.9$\pm$0.7	& 38.5$\pm$1.5	& 31.7$\pm$1.3	& 33.2$\pm$0.9	& 37.8$\pm$3.0	& 40.6$\pm$3.6	& 81.1$\pm$1.8	& 37.6$\pm$3.3	& 37.1$\pm$4.0	& 76.7$\pm$2.0	& 44.10	\\
mSDA \cite{Chen12mSDA}	& 45.3$\pm$1.9	& 37.4$\pm$1.2	& 38.3$\pm$0.7	& 40.3$\pm$1.8	& 33.7$\pm$1.3	& 35.5$\pm$1.1	& 38.1$\pm$2.8	& 40.4$\pm$4.0	& 82.0$\pm$1.8	& 38.5$\pm$3.4	& 37.8$\pm$3.7	& 79.0$\pm$2.1	& 45.51 \\
IGLDA \cite{Jiang16IGLDA}	& \textbf{51.0}$\pm$2.7	& 38.4$\pm$1.9	& 38.6$\pm$1.2	& 41.5$\pm$1.8	& 36.4$\pm$2.2	& 34.2$\pm$1.5	& 38.9$\pm$2.5	& 45.1$\pm$2.4	& 82.6$\pm$1.8	& 40.0$\pm$3.4	& 42.2$\pm$3.6	& \textbf{82.4}$\pm$2.4	& 47.61 \\

MIDA	& 50.3$\pm$2.5	& 39.2$\pm$1.9	& 39.8$\pm$1.0	& 42.7$\pm$1.8	& 35.5$\pm$1.1	& 35.7$\pm$0.7	& 42.3$\pm$2.8	& 45.7$\pm$3.6	& 82.2$\pm$2.0	& 42.8$\pm$2.8	& 43.6$\pm$5.0	& \textbf{82.4}$\pm$2.0	& \textbf{48.51} \\
SMIDA	& 50.5$\pm$2.4	& 39.1$\pm$1.8	& 39.8$\pm$1.1	& 42.7$\pm$2.0	& 35.5$\pm$1.2	& 35.4$\pm$0.8	& 42.4$\pm$2.6	& 45.8$\pm$3.3	& 82.5$\pm$2.1	& \textbf{42.9}$\pm$2.8	& 43.4$\pm$5.1	& 81.9$\pm$2.0	& {48.49} \\
		\hline \hline
	\end{tabular}
		\normalsize
\end{table*}

\begin{table*}[]
	\renewcommand{\tabcolsep}{0.1cm}
	\renewcommand{\arraystretch}{1.5}
	\renewcommand{\doublerulesep}{0pt}
	\caption{Semi-supervised domain adaptation accuracy (\%) on the visual object recognition dataset. Bold values indicate the best results.}
	\label{tbl:visAccSemisup}
	\centering
	\scriptsize
	\begin{tabular}{l lllll lllll lll}
		\hline \hline
			& C\rar A	& D\rar A	& W\rar A	& A\rar C	& D\rar C	& W\rar C	& A\rar D	& C\rar D	& W\rar D	& A\rar W	& C\rar W	& D\rar W & Average \\	
		\hline
Ori. ft.	& 48.8$\pm$1.8	& 44.5$\pm$1.6	& 43.5$\pm$1.4	& 41.6$\pm$1.9	& 36.7$\pm$2.1	& 37.3$\pm$1.4	& 48.1$\pm$4.2	& 49.3$\pm$3.2	& 81.7$\pm$2.4	& 51.0$\pm$3.4	& 50.9$\pm$4.4	& 80.5$\pm$2.2	& 51.17 \\
KPCA	& 53.3$\pm$2.4	& 46.2$\pm$1.7	& 43.2$\pm$1.1	& 44.1$\pm$1.4	& 39.1$\pm$1.8	& 37.8$\pm$1.1	& 47.3$\pm$3.3	& 53.9$\pm$3.4	& 81.5$\pm$2.9	& 49.7$\pm$2.7	& 54.0$\pm$4.1	& 81.1$\pm$2.2	& 52.60 \\
TCA \cite{Pan11TCA}	& \textbf{55.3}$\pm$2.2	& \textbf{48.6}$\pm$1.8	& 45.7$\pm$1.4	& \textbf{46.1}$\pm$2.0	& 40.3$\pm$2.1	& \textbf{39.7}$\pm$1.4	& 52.1$\pm$3.0	& 56.3$\pm$4.5	& \textbf{83.7}$\pm$2.9	& 55.4$\pm$3.5	& 58.3$\pm$4.5	& 84.2$\pm$1.9	& {55.46} \\
SSTCA \cite{Pan11TCA}	& 55.3$\pm$2.2	& \textbf{48.6}$\pm$1.8	& 45.6$\pm$1.4	& 46.0$\pm$2.0	& 40.3$\pm$2.1	& \textbf{39.7}$\pm$1.3	& 52.1$\pm$3.1	& 56.3$\pm$4.6	& \textbf{83.7}$\pm$2.9	& 55.4$\pm$3.5	& 58.4$\pm$4.5	& 84.2$\pm$1.9	& {55.47} \\
ITL \cite{Shi12Itl}	& 51.5$\pm$3.1	& 47.7$\pm$2.5	& 44.1$\pm$2.1	& 40.0$\pm$2.2	& 36.8$\pm$3.2	& 36.6$\pm$2.2	& 44.4$\pm$4.1	& 48.2$\pm$4.0	& 59.7$\pm$2.6	& 51.5$\pm$4.5	& 54.9$\pm$3.8	& 68.5$\pm$3.3	& 48.65 \\
SA \cite{Fern13Sa}	& 51.8$\pm$2.3	& 47.6$\pm$2.8	& 44.7$\pm$1.7	& 42.6$\pm$1.7	& 36.9$\pm$2.9	& 36.8$\pm$2.0	& 45.3$\pm$4.2	& 49.0$\pm$3.9	& 71.0$\pm$2.9	& 47.2$\pm$2.9	& 49.0$\pm$3.6	& 76.1$\pm$2.5	& 49.82 \\
GFK \cite{Gong14Gfk}	& 46.1$\pm$0.6	& 46.2$\pm$0.6	& 46.2$\pm$0.7	& 39.6$\pm$0.4	& 33.9$\pm$0.6	& 32.3$\pm$0.6	& 50.9$\pm$0.9	& 55.0$\pm$0.9	& 74.1$\pm$0.9	& 56.9$\pm$1.0	& 57.0$\pm$0.9	& 80.2$\pm$0.4	& 51.53 \\
LTSL \cite{Shao14Ltsl}	& 50.4$\pm$0.5	& 47.4$\pm$0.5	& \textbf{47.8}$\pm$0.4	& 39.8$\pm$0.4	& 36.7$\pm$0.4	& 38.5$\pm$0.3	& \textbf{59.1}$\pm$0.7	& \textbf{59.6}$\pm$0.6	& 82.6$\pm$0.5	& \textbf{59.5}$\pm$1.1	& \textbf{59.5}$\pm$0.8	& 78.3$\pm$0.4	& 54.93 \\
SCL \cite{Blit07SCLMI}	& 48.8$\pm$1.7	& 45.0$\pm$1.4	& 43.4$\pm$1.3	& 41.3$\pm$1.8	& 36.3$\pm$2.2	& 37.4$\pm$1.4	& 48.9$\pm$4.4	& 49.3$\pm$3.5	& 81.8$\pm$2.5	& 51.9$\pm$3.7	& 52.0$\pm$4.4	& 81.0$\pm$2.2	& 51.42 \\
mSDA \cite{Chen12mSDA}	& 50.4$\pm$2.2	& 48.1$\pm$1.6	& 45.6$\pm$1.6	& 43.6$\pm$1.9	& 38.9$\pm$2.2	& 39.3$\pm$1.5	& 48.9$\pm$4.5	& 49.2$\pm$4.9	& 82.3$\pm$2.5	& 52.9$\pm$3.9	& 52.1$\pm$4.4	& 81.6$\pm$2.1	& 52.75 \\

MIDA	& 55.2$\pm$2.2	& \textbf{48.6}$\pm$1.8	& 45.6$\pm$1.4	& \textbf{46.1}$\pm$2.0	& \textbf{40.4}$\pm$2.2	& \textbf{39.7}$\pm$1.4	& 52.1$\pm$3.0	& 56.4$\pm$4.6	& \textbf{83.7}$\pm$2.8	& 55.3$\pm$3.4	& 58.5$\pm$4.6	& 84.2$\pm$1.8	& {55.48} \\
SMIDA	& 55.2$\pm$2.2	& \textbf{48.6}$\pm$1.8	& 45.7$\pm$1.4	& \textbf{46.1}$\pm$2.0	& 40.3$\pm$2.1	& \textbf{39.7}$\pm$1.4	& 52.2$\pm$3.1	& 56.3$\pm$4.6	& \textbf{83.7}$\pm$2.9	& 55.4$\pm$3.5	& 58.5$\pm$4.5	& \textbf{84.3}$\pm$1.9	& \textbf{55.49} \\
		\hline \hline
	\end{tabular}
		\normalsize
\end{table*}

For GFK, low-rank transfer subspace learning (LTSL), domain adaptation by shifting covariance (DASC), and a recent method called integration of global and local metrics for domain adaptation (IGLDA), we copied the best results reported in the original papers \cite{Gong14Gfk}, \cite{Shao14Ltsl}, \cite{Cui14Flow}, \cite{Jiang16IGLDA}. For other methods tested, the hyper-parameters were tuned for the best accuracy. Logistic regression was adopted as the classifier. The polynomial kernel with degree 2 was used in KPCA, TCA, SSTCA, MIDA, and SMIDA.  The domain features were defined according to the domain labels using the one-hot coding scheme. MIDA and SMIDA achieve the best average accuracies in both unsupervised and semi-supervised visual object recognition experiments. We observe that TCA and SSTCA have comparable performance with MIDA and SMIDA, which may be explained by the fact that the HSIC criterion used in MIDA and MMD used in TCA are identical under certain conditions when there are one source and one target domain \cite{Song12FtSelHsic}. Besides, the feature augmentation strategy in MIDA is not crucial in this dataset because there is no change in conditional probability. On the other hand, TCA and SSTCA can only handle one source and one target domains. SSTCA uses the manifold regularization strategy to preserve local geometry information, hence introduces three more hyper-parameters than SMIDA. Moreover, computing the data adjacency graph in SSTCA and the matrix inversion operation in TCA and SSTCA make them slower than MIDA and SMIDA. We compared their speed on the domain adaptation experiment C\rar A. They were run on a server with Intel Xeon 2.00 GHz CPU and 128 GB RAM. No parallel computing was used. The codes of the algorithms were written in Matlab R2014a. On average, the running times of each trial of MIDA, SMIDA, TCA, and SSTCA were 2.4 s, 2.5 s, 3.0 s, and 10.2 s, respectively. Therefore, MIDA and SMIDA are more practical to use than TCA and SSTCA. Besides, they were initially designed for drift correction. This dataset is used to show their universality.

\section{Conclusion} \label{sec:conclusion}

In this paper, we introduced maximum independence domain adaptation (MIDA) to learn domain-invariant features. The main idea of MIDA is to reduce the inter-domain discrepancy by maximizing the independence between the learned features and the domain features of the samples. The domain features describe the background information of each sample, such as the domain label in traditional domain adaptation problems. In the field of sensors and measurement, the device label and acquisition time of the each collected sample can be expressed by the domain features, so that unsupervised drift correction can be achieved by using MIDA. The feature augmentation strategy proposed in this paper adds domain-specific biases to the learned features, which helps MIDA to align domains.

MIDA and SMIDA are flexible algorithms. With the design of the domain features and the use of the HSIC criterion, they can be applied in all kinds of domain adaptation problems, including discrete or continuous distributional change, supervised/semi-supervised/unsupervised, multiple domains, classification or regression, etc. They are also easy to implement and fast, requiring to solve only one eigenvalue decomposition problem. Future directions may include further extending the definition of the domain features for other applications.

\ifCLASSOPTIONcaptionsoff
  \newpage
\fi

\bibliographystyle{IEEEtran}

\bibliography{MIDA}

\end{document}